# Internet of Things: Digital Footprints Carry A Device Identity

Rajarshi Roy Chowdhury[1, 2, a)], Azam Che Idris[1] and Pg Emeroylariffion Abas[1]

[1]Faculty of Integrated Technologies, Universiti Brunei Darussalam, Jalan Tungku Link, Gadong BE1410, Brunei Darussalam
[2]Department of Computer Science and Engineering, Sylhet International University, Shamimabad Road, Sylhet 3100, Bangladesh

Corresponding author: [a)] 19h0901@ubd.edu.bn or rajarshiry@gmail.com

**ABSTRACT.** The usage of technologically advanced devices has seen a boom in many domains, including education, automation, and healthcare; with most of the services requiring Internet-connectivity. To secure a network, device identification plays key role. In this paper, a device fingerprinting (DFP) model, which is able to distinguish between Internet of Things (IoT) and non-IoT devices, as well as uniquely identify individual devices, has been proposed. Four statistical features have been extracted from the consecutive five device-originated packets, to generate individual device fingerprints. The method has been evaluated using the Random Forest (RF) classifier and different datasets. Experimental results have shown that the proposed method achieves up to 99.8% accuracy in distinguishing between IoT and non-IoT devices and over 97.6% in classifying individual devices. These signify that the proposed method is useful in assisting operators in making their networks more secure and robust to security breaches and unauthorised access.

**Keywords :** *digital footprint; network traffic traces; machine learning algorithm; internet of things; device fingerprinting*

## INTRODUCTION

It has been predicted that the number of network-connected Internet of Things (IoT) and non-IoT devices worldwide will reach approximately 30.9 billion and 10.3 billion, respectively, by the year 2025 [1]. Proliferated growth of these devices with their heterogeneous functionalities, has imposed new challenges to network administrators and operators, in providing, managing, and controlling the operations and security of the network services [2]. Accurate device identification is one key aspect that needs to be seriously considered in securing network-connected devices. Conventionally, internet protocol (IP) enabled devices have been using user-defined identifiers, such as IP and media access control (MAC) addresses, as a form of identifications. However, these identifiers have been proven to be vulnerable [3] to various attacks, such as spoofing [4] and device mobility, due to the availability of malicious software [5], for performing such attacks. Device fingerprinting (DFP) [3] represents one technique that may be used to identify devices based on their communication traffic traces (or digital footprints) without using explicit identifiers, and it can be performed, either actively or passively, from different layers of the communication model [6].

Due to the prominent characteristics of network traffic features, many researchers [2, 7] have used packet-level features for different purposes [8], including for device identification [9]. Sivanathan et al. [10] have described a DFP scheme based on the analysis of passively observed network traffic traces. A total of 11 statistical features are used as device fingerprints, from packet traffic-flows over a period of one day, by looking at the devices' sleeping time, average packet size and traffic rate, active time, number of servers and protocols used in a flow, number of unique domain name system (DNS) request, and intervals of DNS and network time protocol (NTP) requests. Subsequently, these features are used to train an ML model for classification. It has been shown that the DFP scheme is able to distinguish between IoT and non-IoT devices with high accuracy and achieve over 95% accuracy in identifying individual IoT devices. The same researchers [9] have also presented another device fingerprinting scheme, by utilizing statistical characteristics of hourly network traffic traces, to generate 8 device-specific fingerprints. Experimental result has shown over 99% accuracy using the UNSW dataset. Charyyev et al. [11] have utilized Nilsimsa hash value of packet flows (*n* packets) for device-specific fingerprints, to classify individual IoT devices, to achieve 93% precision.

Researchers in [2, 12] have used 12 packets information, to generate device signatures for classifying IoT devices, with 81.5% global accuracy and 76.15% accuracy using an aggregated model, whilst Aksoy and Gunes [13] have presented a DFP approach, known as SysID, which utilizes features from a single packet, for identifying smart home IoT devices with 82% average classification accuracy. Bezawada et al. [14] have utilized 5 consecutive packets information, including protocols headers and payload (20 features), for classifying IoT devices uniquely with mean identification accuracy of 93% to 100% using a laboratory dataset of 14 IoT devices. In [15], the authors have used a one second window to group packets, for generating statistical fingerprinting features. These features are then used to train a binary classifier for categorizing IoT and non-IoT devices with high accuracy of 99%, whilst a multi-class classifier has been used to uniquely identify IoT devices with about 96% accuracy. All these existing DFP models, however, require either a large number of features set from different layers of the communication model, or a large number of network packets information for generating fingerprints. Consequently, these models consume a long period of time, and require complex computation. As such, a more efficient DFP model is required for classifying devices with high accuracy, but with less computation cost.

In this paper, a supervised machine learning (ML) based DFP model, which generates device-specific signatures by computing four statistical features from consecutive five packets of the network traffic, has been proposed. An intuition that these features carry device-specific characteristics in terms of device memory and processing speed. Experimental results have shown that over 97.0% accuracy is achievable in classifying individual non-IoT devices from traffic collected in a laboratory environment, and 97.3% accuracy on the non-IoT traffic traces from the UNSW dataset. The proposed DFP model is also capable of distinguishing between IoT and non-IoT devices with up to 99.8% accuracy on the UNSW dataset. The key contributions of this research work are:

- Identifying device-specific features from the device-originated communication traffic traces, to generate device signatures for classification.
- Instrument an experimental testbed of non-IoT devices in a laboratory environment for data collection.
- Evaluate the proposed DFP scheme performance based on a supervised ML algorithm, to distinguish between IoT and non-IoT devices and identify individual devices.

The rest of the paper is organized as follows. The proposed ML-based device fingerprinting method, as well as the datasets, data collection procedure, and an ML classifier are described in Section II. Section III describes experimental results on various datasets, and finally, conclusion is given in Section IV.

## METHODOLOGY

The proposed DFP method is used to extract unique device features from network traffic traces. These features are used to train an ML classifier, and subsequently, used to test the performance of the proposed DFP method on different datasets. This section describes the proposed DFP method, the datasets used for training and testing, as well as the classification method used to test the model.

### Datasets: IoT and Non-IoT

The proposed device fingerprinting model performance has been evaluated by utilizing a publicly available dataset: UNSW [9], and a testbed dataset of non-IoT devices, which has been collected from a laboratory environment. Summary of the datasets are listed in Table 1. The UNSW dataset comprises network traffic traces from both IoT and non-IoT devices, including TP-Link camera, smart bulb, Belkin camera, smart doorbell, printer, smart photo frame, laptop, smartphone, and tablet devices, with these heterogeneous devices coming from different manufacturers: Belkin, Philips Hue, Netatmo, TP-Link, Withings, HP, Apple. On the other hand, the laboratory dataset comprises 7 non-IoT devices, including laptops, smartphones, and desktops, from different manufacturers. The data collection procedure from the 7 non-IoT devices is described in the following section.

**TABLE 1.** List of IoT and non-IoT Datasets.

| Dataset | Devices | | Total Packets | Source |
| --- | --- | --- | --- | --- |
| | IoT | Non-IoT | | |
| UNSW | 22 | -- | 6,844,740 | [9] |
| | -- | 7 | 3,515,705 | |
| Lab Dataset | -- | 7 | 442,970 | -- |

**TABLE 2.** List of non-IoT devices for experimental set up.

| No. | Device Category | Device Name/Model | Operating System | Connectivity | MAC Address |
| --- | --- | --- | --- | --- | --- |
| 1 | Laptop | Aspire-S7 | Windows | WiFi | 34:23:87:b7:56:17 |
| 2 | | ProBook-4410s | | WiFi/Ethernet | 00:25:b3:47:da:6f |
| 3 | Desktop | Asus | | Ethernet | 08:60:6e:c1:79:c2 |
| 4 | | HP-EliteDesk | | Ethernet | 80:e8:2c:d6:9e:49 |
| 5 | Smart Phone | MYA-U29 | Android | WiFi | d0:ff:98:95:57:af |
| 6 | | MLXP2ZA-A | iOS | WiFi | e0:c7:67:45:a3:62 |
| 7 | | MWC22KH-A | | WiFi | 06:44:b7:aa:20:98 |

## Dataset Collection Methodology

An experimental design, consisting of local area network (LAN) and wireless local area network (WLAN) with non-IoT devices, was set up in a laboratory environment at Universiti Brunei Darussalam (UBD). Design of the testbed is depicted in Figure 1, with the seven non-IoT devices from different manufacturers and of different types, as listed in Table 2. These devices were configured, to connect with an access point (AP) either using ethernet or wireless fidelity (WiFi) interfaces.

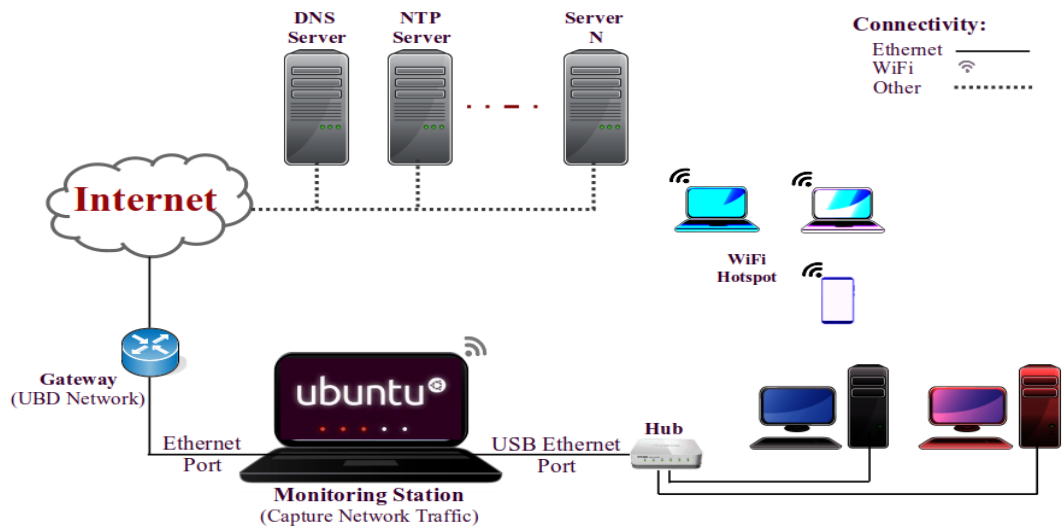

**FIGURE 1.** An experimental testbed of non-IoT devices network (LAN/WLAN).

A laptop was used to configure an access point (AP), which was used to provide network services to the non-IoT devices, as well as to monitor and capture communication footprints from the devices. The *Dell Inspiron 15 5000 Series* laptop runs Ubuntu 18.04 as an operating system (OS), and was connected to the UBD network via its built-in Ethernet interface, to provide the Internet connections. The built-in WiFi interface was configured as a *WiFi Hotspot*, providing wireless connectivity to the WiFi-enabled (IEEE 802.11 standard) devices. Additionally, a *TU3-ETG* USB Ethernet adapter was connected to the laptop, and used to set up a LAN network using the *D-Link Switch Hub DES-1005A* hub for providing network services to the connected non-IoT devices. On the Ubuntu OS, the network connection editor tool, i.e. *nm-connection-editor*, was been utilised for connection establishment.

Devices generally generate two types of traffic [9]: autonomous traffic, including traffic generated for connection establishment, application and system synchronizations, and activity traffic, which is generated due to human or object interactions. These inbound and outbound communication traffic traces, flowing over both interfaces (external Ethernet and built-in WiFi interfaces) were captured using *tcpdump* 4.9.3 utility, and stored into .pcap (packet capture) files format, similar to [16]. Device-originated traffic traces were then extracted using *TShark* utility and stored in .csv (comma-separated values) files format, along with labelling of individual devices names. Finally, the recorded dataset was cleaned for further processing, by eliminating inconsistent instances, including empty rows, and duplicate values.

## Device Fingerprinting Model

The proposed DFP scheme architecture is depicted in Figure 2, which uses device-originated communication traffic traces to generate device fingerprints for classification. Device-originated traffic traces are filtered according to individual device MAC addresses, with *tcp.window_size* and *ip.len* values extracted from each packet from the available captured data. These two values of a network packet carry significant device-specific information. *tcp.window_size* value depends on a device buffer size and computation speed [14] whilst *ip.len* value specifies the total length of a packet to represent unique characteristics of a devices communication pattern [15]. *tcp.window_size* and *ip.len* values from five consecutive packets (as one instance) are utilized, to compute mean ($\mu$) and standard deviation ($\sigma$), for constructing device-specific fingerprints, i.e. *iplen_$\mu$*, *iplen_$\sigma$*, *tcpwinsiz_$\mu$*, and *tcpwinsiz_$\sigma$*. These 4 statistical fingerprints have been used for training a machine learning (ML) model, and subsequently, to evaluate the performance, of the model in classifying devices using datasets, which have been randomly split into training (80% instances) and testing (20% instances) datasets.

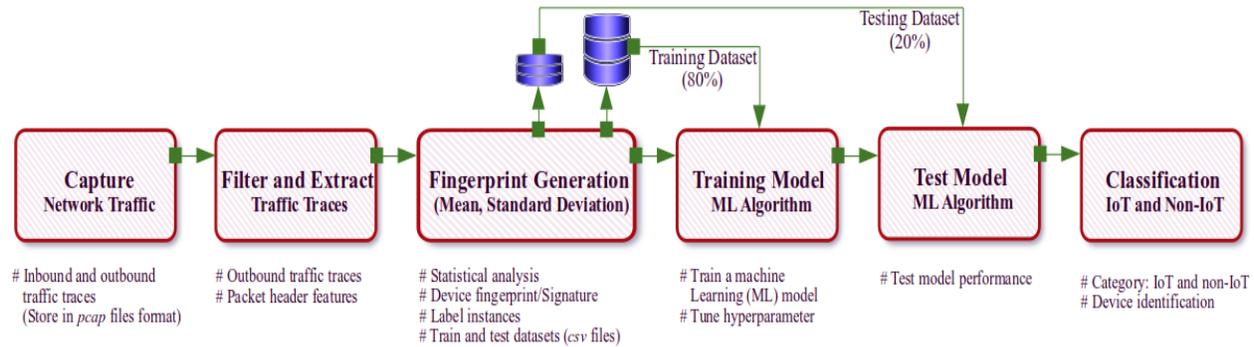

**FIGURE 2.** The proposed device fingerprinting scheme.

## Random Forest Classifier

Random Forest (RF) classifier is a supervised machine learning (ML) algorithm, that can be used for both classification [9] and regression [17] problems. The algorithm randomly generates a group of trees, with majority voting used to make a decision from the ensemble of decision trees [18, 19], for the classification task, as presented in Figure 3. This assists in avoiding over-fitting problem. Researchers in different domains have utilized RF classifier for different classification tasks. In [9], the RF algorithm has been used for classifying IoT devices with high accuracy. Primartha et al. [20] have performed anomaly detection using the algorithm, and it has also been used for disease identification in medical science [21]. In this paper, RF classifier is used to appraise the performance of the proposed DFP method, by using the extracted features for training the RF classifier, and subsequently, using the trained RF classifier to determine classification performance. Some of the significant tunable hyper-parameters are set experimentally, including the number of iterations (or number of trees) = 100, seed = 1, and batch size (number of instances) = 100, to improve classification accuracy and reduce the root mean squared error (RMSE) [22].

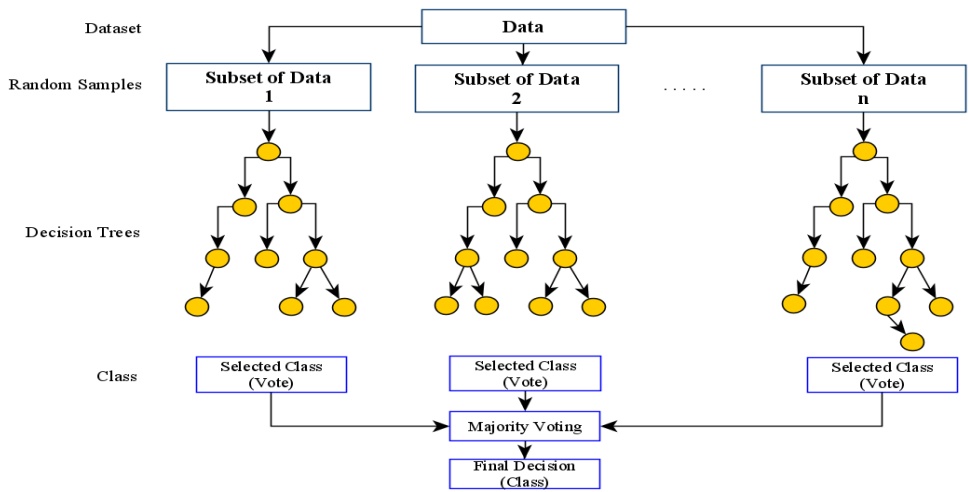

**FIGURE 3.** An abstract representation of a RF classifier.

## RESULTS AND DISCUSSION

The proposed DFP method has been evaluated using *waikato environment for knowledge analysis* (Weka) tool [23]. An online dataset: UNSW [9] dataset, and an experimental dataset, as presented in Table 3, have been utilized to evaluate the classification performance based on the RF classifier. The UNSW dataset consists of network traffic traces from IoT and non-IoT devices, which are referred to as the U-IoT and U-NonIoT datasets, respectively. On the other hand, the experimental dataset contains only network traffic traces from non-IoT devices, and it is referred to as the L-NonIoT dataset.

**TABLE 3.** Total number of instances used for evaluating the proposed DFP model.

| Dataset | Devices | | Training Dataset (80%) | Test Dataset (20%) | Total Instances (100%) |
|---|---|---|---|---|---|
| | IoT | Non-IoT | | | |
| UNSW (U-IoT) | * | --- | 1,095,158 | 273,790 | 1,368,948 |
| UNSW (U-NonIoT) | --- | * | 562,513 | 140,628 | 703,141 |
| Lab (L-NonIoT) | --- | * | 70,875 | 17,719 | 88,594 |

The proposed DFP method utilises 5 network traffic packets as one instance to generate fingerprint. As such, a total of 1,368,948 (6,844,740 / 5) and 703,141 (3,515,705 / 5) instances have been used from the U-IoT and U-NonIoT datasets, respectively, whilst a total of 88,594 (442,970 / 5) instances have been used from the L-NonIoT dataset. 80% of the datasets have been used for training and the remainder for testing. The performance of the trained RF classifier has been measured with respect to its ability to a) distinguish between IoT and non-IoT devices, and b) classify individual devices.

## Device Category: IoT and Non-IoT Devices

Classification performances of the proposed DFP model in distinguishing between IoT and non-IoT devices are presented in Figure 4, on combined U-IoT and U-NonIoT datasets (i.e. UNSW dataset), and combined U-IoT and L-NonIoT datasets. The figure shows that device categorization accuracy reaches up to 99.9% using the RF classifier on the combined U-IoT and L-NonIoT datasets. On the UNSW dataset [9], which consists of instances from 22 IoT and 7 non-IoT devices, the proposed DFP method achieves 99.8% accuracy.

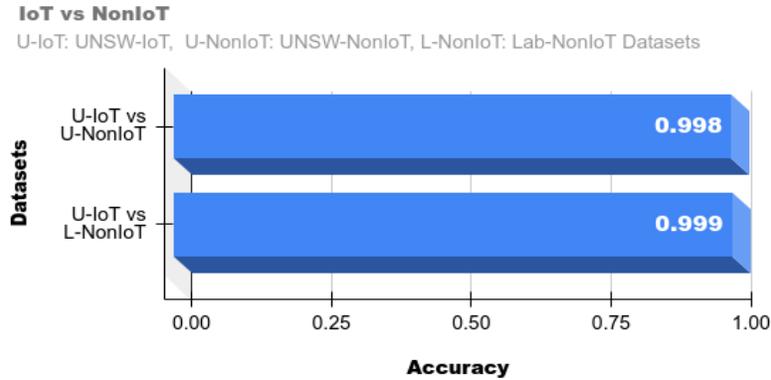

**FIGURE 4.** Categorize IoT and non-IoT devices: UNSW and Lab datasets.

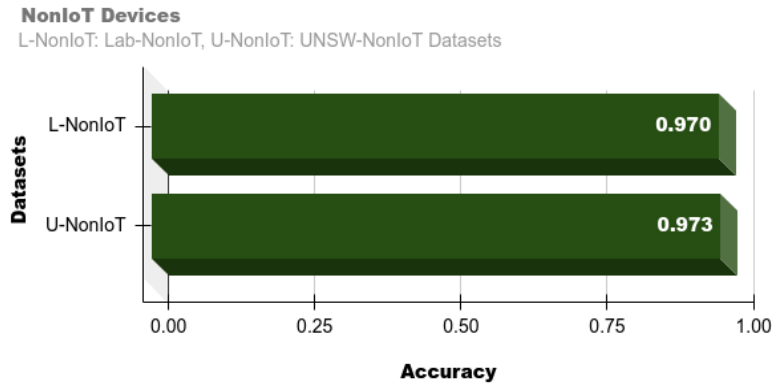

**FIGURE 5.** Classification performance of the non-IoT devices: UNSW and Lab datasets.

## Individual Device Classification

The performances of the proposed DFP method in classifying individual IoT and non-IoT devices on different datasets, are depicted in Figure 5 and Figure 6. In Figure 5, the proposed DFP model achieves over 97.0% accuracy in classifying non-IoT devices from the L-NonIoT and U-NonIoT datasets, with accuracy a little bit higher on the U-NonIoT dataset. Individual IoT devices classification performance of the proposed DFP model, on the U-IoT dataset with 22 IoT devices, is given in Figure 6. Most of the IoT devices in the dataset can be classified with over 97.6% accuracy, with the exception of the BlipcareBPmeter, the BelkinWemoSensor and BelkinWemoSwitch devices, which give classification accuracies of about 75.0%, 96.5% and 91.4%, respectively. The lowest accuracy for the BlipcareBPmeter device is due to the limited number of instances available from this device for training and testing.

## CONCLUSION

A large number of heterogeneous IoT and non-IoT devices from different manufacturers are being connected to the Internet, to obtain network-based services. In terms of network security, it is challenging for network administrators and operators to identify the connected devices using conventional identifiers, as they are prone to security breaches. In this paper, a DFP model based on the analysis of network traffic traces has been proposed, which is capable of distinguishing between IoT and non-IoT devices as well as classifying individual IoT and non-IoT devices. As opposed to other methods in the literature, which require relatively large number of features and requiring longer sequence of packet network traffics to construct their DFP features, only 4 statistical features from 5 consecutive packet network traffics are required to construct the DFP features. These are used for training and testing an ML classifier. Evaluations on the UNSW dataset have shown that the proposed DFP method is able to distinguish between IoT and non-IoT devices with up to 99.8% accuracy, and individually classify most of the IoT and non-IoT devices with over 97.6% accuracy. On the laboratory collected network traces, the proposed DFP model is able to classify individual devices with 97.0% accuracy. The research outcomes signify that the proposed DFP model is useful for device identification and may assist network administrators in providing a more secure network.

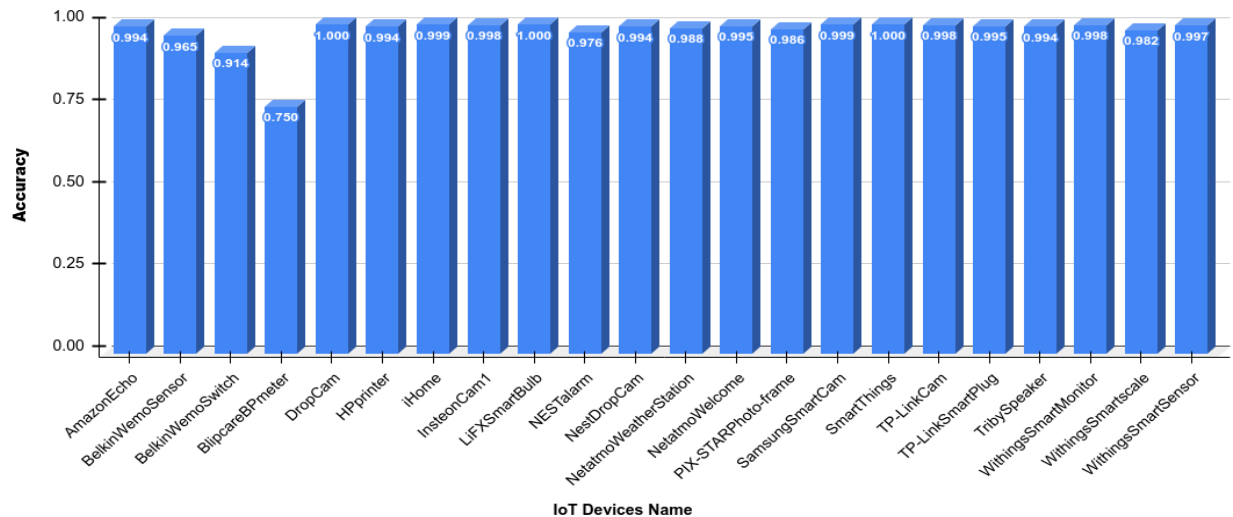

**FIGURE 6.** Individual IoT device classification performance: U-IoT dataset.

## ACKNOWLEDGEMENTS

The authors are profoundly grateful to the Faculty of Integrated Technologies (FIT), Universiti Brunei Darussalam (UBD), for supporting this research work, as well as to UBD for awarding the UBD Graduate Scholarship (UGS) to the first author.